\documentclass[11pt,a4paper]{article}

\usepackage[dvipsnames]{xcolor}
\usepackage{tikz,pgfplots}
\usepackage{subfigure}

\usepackage{times,latexsym}
\usepackage{url}
\usepackage[T1]{fontenc}

\usepackage{amsmath}
\usepackage{multirow}

\usepackage{graphicx}
\usepackage{rotating}
\usepackage{gb4e} 
\noautomath
\usepackage{todonotes}
\usepackage{eurosym}
\usepackage{subfigure}
\usepackage{textcomp}
\usepackage{soul}
\usepackage{color}

\usepackage{tikz-dependency}
\usepackage{standalone}

\usepackage[acceptedWithA]{tacl2018v2}
\newcommand{\emp}{$\rule{1em}{0.15mm}$~}
\newcommand{\ignore}[1]{}

\DeclareMathSymbol{\mlq}{\mathord}{operators}{``}
\DeclareMathSymbol{\mrq}{\mathord}{operators}{`'}

\definecolor{choice_strong}{RGB}{220,80,80}
\definecolor{choice_medium}{RGB}{215,150,140}
\definecolor{choice_weak}{RGB}{250,220,220}

\usepackage[]{tacl2018v2}

\title{Where's My Head? Definition, Dataset and Models \\for Numeric Fused-Heads Identification
and Resolution}

\author{Yanai Elazar$^\dagger$ \and Yoav Goldberg$^\dagger$ $^\ast$ \\[0.5em]
$^\dagger$Computer Science Department,  Bar-Ilan University, Israel \\[0.5em]
$^\ast$Allen Institute for Artificial Intelligence \\[0.5em]
\texttt{\{yanaiela,yoav.goldberg\}@gmail.com} 
}

\date{}

\begin{document}
\maketitle
\begin{abstract}
We provide the first computational treatment of fused-heads constructions (FH),
focusing on the numeric fused-heads (NFH).
FHs constructions are noun phrases (NPs) in which the head noun is missing and is said
to be ``fused'' with its dependent modifier. This missing information is implicit and is important for sentence understanding.
The missing references are
easily filled in by humans but pose a challenge for computational models.
We formulate the handling of FH as a two stages process: \textbf{identification} of the FH construction and \textbf{resolution} of the missing head. We explore
the NFH phenomena in large corpora of English text and create (1) a dataset and
a highly accurate method for NFH identification; (2) a 10k examples (1M tokens)
crowd-sourced dataset of NFH resolution;  and (3) a neural baseline for the NFH
resolution task. We release our code and dataset, in hope to foster further
research into this challenging problem.
\end{abstract}

\section{Introduction}
Many elements in language are not stated explicitly but need to be inferred from
the text. This is especially true in spoken language but also holds for written
text. Identifying the missing information and filling the gap is a crucial part
of language understanding. 
Consider the sentences below:

\begin{exe}
	\ex\label{ex1} I'm \textbf{42} \emp, Cercie.
    \ex\label{ex2} It's worth about \textbf{two million} \emp.
    \ex\label{ex3} I've got two \textit{months} left, \textbf{three} \emp at the most.
	\ex\label{ex4} I make an amazing \textit{Chicken Cordon Bleu}.\\
	She said she'd never had \textbf{one}
\end{exe}

In (\ref{ex1}), it is clear that the sentence refers to the \emph{age} of the speaker, but
this is not stated explicitly in the sentence. Similarly, in (\ref{ex2}) the
speaker discusses the worth of an object in some \emph{currency}. In (\ref{ex3}), the number refers back to an object already mentioned before---\emph{months}. 

All of the above examples are of \emph{numeric fused heads} (NFH), a linguistic
construction which is a subclass of the more general \emph{fused heads} (FH)
construction, limited to numbers.  
FH are noun phrases (NPs) in which the head noun is missing and is said
to be ``fused'' with its dependent modifier \cite{huddleston2002cambridge}. In the examples above, the numbers `\emph{42}', `\emph{two million}', `\emph{three}' and
`\emph{one}' function as FHs, whereas their actual heads (\textsc{years old}, \textsc{dollar}, \emph{months}, \emph{Chicken Cordon Bleu}) are missing and need to be inferred.

While we focus on NFH, FH in general can occur also with other
categories, such as determiners and adjectives.
For example in the following sentences:
\begin{exe}
	\ex\label{fh_ex1} Only the \textbf{rich} \emp will benefit.
	\ex\label{fh_ex2} I need some \textit{screws} but can't find \textbf{any} \emp.
\end{exe}
the adjective `\emph{rich}' refers to rich \textsc{People} and the determiner `\emph{any}' refers to \emph{screws}. In this work we focus on the \textit{numeric fused head}.
\begin{table*}[t]
\begin{center}
\resizebox{\textwidth}{!}{
\begin{tabular}{|l|l|l|}
\hline
Index & Text & Missing Head \\ \hline \hline
i & Maybe I can teach the kid a \textit{thing} or \textbf{two} \emp. & \textit{thing} \\ \hline
ii & you see like \textbf{3} \emp or 4 \textit{brothers} talkin' & \textit{brothers} \\ \hline
iii & When the clock strikes \textbf{one}... the Ghost of Christmas Past
 & \textsc{O'clock} \\ \hline
iv & My manager says I'm a perfect \textbf{10}! & \textsc{Score} \\ \hline
v & See, that's \textbf{one} \emp of the \textit{reasons} I love you & \textit{reasons} \\ \hline
vi & Are you \textbf{two} done with that helium? & \textsc{People}\\ \hline
vii & No \textbf{one} cares, dear. & \textsc{People} \\ \hline
viii & \textit{Men} are like \textit{busses}: If you miss \textbf{one} \emp, you can be sure there'll be soon another \textbf{one} \emp... & \textit{Men} $|$ \textit{busses} \\ \hline
ix & I'd like to wish a happy \textbf{1969} to our new President. & \textsc{Year} \\ \hline
x & I probably feel worse than Demi Moore did when she turned \textbf{50}. & \textsc{Age} \\ \hline
xi & How much was it? \textbf{Two hundred}, but I'll tell him it's fifty. He doesn't care about the gift; & \textsc{Currency} \\ \hline
xii & Have you ever had an \textit{unexpressed thought}? I'm having \textbf{one} \emp now. & \textit{unexpressed thought} \\ \hline
xiii & It's a curious thing, the death of a loved \textbf{one}. & \textsc{People} \\ \hline
xiv & I've taken \textbf{two} \emp over. Some \textit{fussy old maid} and some \textit{flashy young man}. & \textit{fussy old maid} \& \textit{flahy young man} \\ \hline \hline

xv & [non-NFH] \textbf{One} \textit{thing} to be said about traveling by stage. & - \\ \hline
xvi & [non-NFH] After \textbf{seven} long \textit{years}... & - \\ \hline
\end{tabular}
}
\end{center}

\caption{Examples of NFH. The \textit{anchors} are marked in bold, the heads are marked in italic. The missing heads in the last column are written in italic for \textit{Reference} cases and in upper case for the \textit{Implicit} cases. The last two rows contain examples with regular numbers -- which are not considered as NFHs.}
\label{tbl:examples}
\end{table*}

Such sentences often arise in dialog situations as well as other genres.
Numeric expressions play an important role in various tasks, including textual entailment \cite{lev2004solving,dagan2013recognizing}, solving arithmetic problems \cite{roy2015solving}, numeric reasoning \cite{roy2015reasoning,trask2018neural} and language modeling \cite{spithourakis2018numeracy}.

While the inferences required NFH construction may seem trivial for a human hearer, they 
are for the most part not explicitly addressed by current natural language
processing systems. Indeed, tasks such as information extraction, machine translation,
question answering and others could greatly benefit from recovering such
implicit knowledge prior to (or in conjunction with) running the
model.\footnote{To give an example from information extraction, consider a system based on syntactic patterns that needs to handle the sentence ``Carnival is expanding its ships business, with 12 to start operating next July.''. In the context of MT, Google Translate currently translates the English sentence ``I'm in the center lane, going about 60, and I have no choice'' into French as ``Je suis dans la voie du centre, environ \underline{60 ans}, et je n'ai pas le choix'', changing the implicit \emph{speed} to an explicit \emph{time period}.}

We find NFHs particularly interesting to model: they are common (Section \ref{sec:nfh}), easy to understand and resolve by humans (Section \ref{nfh-res-corpus}), important for language understanding, not handled by current systems (Section \ref{sec:related}) and hard for current methods to resolve (Section \ref{sec:nfh-res}).

The main contributions of this work are:
\begin{itemize}
    \item We provide an account of NFH constructions and their distribution in a large corpus of English dialogues, where they account for 41.2\% of the numbers. We similarly quantify the prevalence of NFH in other textual genres, showing that they account for between 22.2\% and 37.5\% of the mentioned numbers.
    \item We formulate the FH \emph{identification} (identifying cases that need to be resolved) and \emph{resolution} (inferring the missing head) tasks.
    \item We create an annotated corpus for NFH \emph{identification} and show that the task can be automatically solved with high accuracy.
    \item We create a 900,000 tokens annotated corpus for NFH \emph{resolution}, comprising of \texttildelow 10K NFH examples, and present a strong baseline model for tackling the \emph{resolution} task.
\end{itemize}

\section{Numeric Fused-Heads}
\label{sec:nfh}

Throughout the paper, we refer to the visible number in the FH as the \textit{anchor} and to the missing
head as the \textit{head}. 

\begin{figure*}[t!]
    \centering
    \includegraphics[scale=1.0]{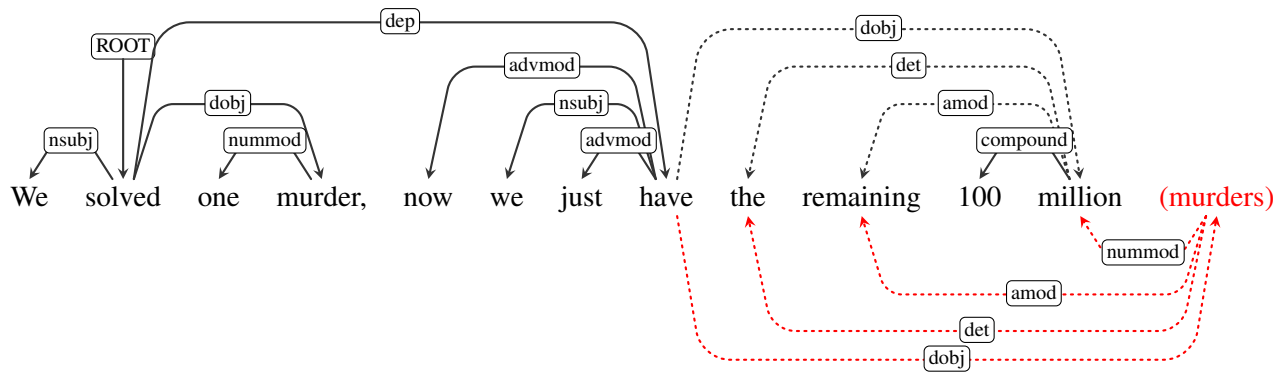}
     \caption{Example for an NFH. The `murders' token is missing, and fused with the `100 million' numeric-span.}
     \label{fig:num-fh}
\end{figure*}

In FH constructions the implicit heads are missing and are said to be \emph{fused} with the anchors, which are either determiners or modifiers. 
In the case of NFH, the modifier role is realized as a number (see examples in Table \ref{tbl:examples}).
The anchors then function both as the determiner/modifier and 
as the head---the parent and the other modifiers of the original head are
syntactically attached to the
anchor. For example, in Figure \ref{fig:num-fh} the phrase \emph{the remaining
100 million} contains an NFH construction with the anchor \emph{100 million},
which is attached to the sentence through the dotted black dependency edges. The
missing head, \emph{murders}, appears in red together with its missing
dependency edges.\footnote{An IE or QA system trying to extract or answer
information about the number of murders being solved will have a much easier time when implicit information would be stated explicitly.}

\paragraph{Distribution}
NFH constructions are very common in dialog situations (indeed, we show in
Section \ref{sec:nfh-iden} that they account for over 40\% of the numbers in a large English corpus of movie dialogs), but are also common in written text such as product reviews or journalistic text.
Using an NFH identification model which we describe in Section \ref{nfh-identification-ml}, we examined the distribution of NFH in different corpora and domains. Specifically, we examined monologues (TED talks \cite{cettoloEtAl:EAMT2012}), Wikipedia (WikiText-2 and WikiText-103 \cite{merity2016pointer}), journalistic text (PTB \cite{marcus1993building}) and product reviews (Amazon reviews\footnote{\url{https://www.kaggle.com/bittlingmayer/amazonreviews}}) in which we found that more than 35.5\%, 33.2\%, 32.9\%, 22.2\% and 37.5\% of the numbers respectively are NFH.



\paragraph{FH Types}
We distinguish between two kinds of FH, which we call \textit{Reference} and \textit{Implicit}. In \emph{Reference} FH, the missing head is referenced explicitly somewhere else in the discourse, either in the same sentence or in surrounding sentences. In \emph{Implicit} FH, the missing head does not appear in the text and needs to be inferred by the reader or hearer based on the context or world knowledge. 

\subsection{FH vs.\ Other Phenomena}
FH constructions are closely related to ellipsis constructions and are also
reminiscent of coreference resolution and other anaphora tasks.

\paragraph{FH vs.\ Ellipsis} With respect to ellipsis, some of the NFH cases we consider can be analyzed as
nominal ellipsis (cf. i, ii in Table \ref{tbl:examples}, and (\ref{ex3}) in the
intro). Other cases of head-less numbers do not traditionally admit an ellipsis
analysis. We do not distinguish between the cases and consider all head-less
number cases as NFH.

\paragraph{FH vs.\ Coreference} With respect to coreference, some
\emph{Reference} FH cases may seem similar to coreference cases. However, we
stress that these are two different phenomena: in coreference, the mention and
its antecedent both refer to the same entity, while the NFH anchor and its
head-reference---like in ellipsis---may share a symbol but do not refer to the
same entity. Existing coreference resolution datasets do consider some FH cases,
but not in a systematic way.
They are also restricted to cases where the antecedent appears in the discourse,
i.e. they do not cover any of the NFH \textit{Implicit} cases.

\paragraph{FH vs.\ Anaphora}
Anaphora is another similar phenomenon. As opposed to coreference, anaphora (and cataphora, which are cases with a forward rather than a backward reference) includes mentions of the same type but different entities. However, the anaphora does not cover our \textit{Implicit} NFH cases, which are not anaphoric but refer to some external context or world knowledge. We note that anaphora/cataphora is a very broad concept, that encompasses many different sub-cases of specific anaphoric relations. There is some overlap between some of these cases and the FH constructions. 



\paragraph{Pronimial \emph{one}} The word \emph{one} is a very common NFH
anchor (61\% of the occurrences in our corpus), and can be used either as a number (viii) or as a pronoun (xiii). The
pronoun usage can be replaced with \emph{someone}. For consistency, we consider
the pronominal usages to be NFH, with the implicit head \textsc{People}.\footnote{While the overwhelming majority of `one' with an implicit \textsc{People} head are indeed pronomial, some cases are not. For example: `\emph{Bailey, if you
don't hate me by now you're a minority of \textbf{one}.}'} 

The \textit{one-anaphora} phenomenon 
was previously studied on its own \cite{gardiner2003identifying,ng2005machine}.
The work by \citet{ng2005machine} divided uses of \emph{one} into six
categories: Numeric (xv),  Partitive (v), Anaphoric (xii),
Generic (vii), Idiomatic (xiii) and Unclassified. We consider all of these, except the Numeric category as NFH constructions.

\ignore{
The FH phenomenon resembles and coincides with two well studied phenomenon: Coreference resolution and Ellipsis. In Coreference, entities only cluster together with the same instances. The \textit{Reference} in the FH phenomenon, might be (and usually is) an Anaphora, meaning it refers to the same class or concept but not necessarily to the same entity.

Ellipsis coincides heavily with the FH phenomena (e.g. Examples 1, 2 in Table \ref{tbl:examples}) but not always (e.g. Examples 3, 4 in Table \ref{tbl:examples}). The major type of studied Ellipsis in the NLP community is the Verb-Phrase Ellipsis. We do not restrict ourselves to this type and allow all types of Ellipsis.

The \textit{One-Anaphora} phenomenon also occurs in our corpus.
The number `one' is the most common token of the corpus (61.0\%) and was previously studied on its own \cite{gardiner2003identifying,ng2005machine}. The work by \citet{ng2005machine} divided uses of one into six categories: Numeric (e.g. ex 1),  Partitive (e.g. ex 6), Anaphoric (e.g. ex 10), Generic (e.g. ex 8), Idiomatic (e.g. ex 9) and Unclassified. All of which might be considered as NFH.
}

\subsection{Inclusive Definition of NFH}
\label{sec:modeling-scope}
While our work is motivated by the linguistic definition of FH, we take a pragramatic approach in which we do not determine the scope of the NFH task based on fine-grained linguistic distinctions. 
Rather, we take an inclusive approach that is motivated by considering the end-user of an NFH resolution system who we imagine is interested in resolving all numbers that are missing a nominal head.
Therefore, we consider all cases that ``look like an NFH'' as NFH, even if the actual linguistic analysis would label them as gapping, ellipsis, anaphoric pronominal-one or other phenomena. We believe this makes the task more consistent and easier to understand to both end users, annotators, and model developers.

\section{Computational Modeling and Underlying Corpus}
\label{sec:nfh-modeling}
We treat the computational handling of FH as two related tasks: identification
and resolution. We create annotated NFH corpora for both.

\paragraph{Underlying Corpus}
\label{sec:corpus}
As the FH phenomenon is prevalent in dialog situation, we base our corpus
on dialog excerpts from movies and TV-series scripts (The IMDB corpus). The corpus contains 117,823 different episodes and movies. Every such item may contain several scenes, with an average of 6.9 scenes per item. Every scene may contain several speaker turns, each of which may span several sentences. The average number of turns per scene is 3.0.
The majority of the scenes have at least two participants. Some of the utterances refer to the global movie context.
\footnote{Referring to a broader context is not restricted to movie-based
dialogues. For example, online product reviews contain examples such as
``\emph{... I had three in total...}'', with \emph{three} referring to the
purchased product, which is not explicitly mentioned in the review.}

\paragraph{NFH Identification}
In the identification stage, we seek NFH anchors within headless NPs which contain a number. More concretely, given a sentence, we seek a list of spans corresponding to all of the anchors within it. An NFH anchor is restricted to a single number, but not a single token. For example, \emph{thirty six} is a two-token number which can serve as an NFH anchor. We assume all anchors are contiguous spans.
The identification task can be reduced to a binary decision, categorizing each numeric span in the sentence as FH/not-FH.

\paragraph{NFH Resolution} The resolution task resolves an NFH anchor to its missing head.
Concretely, given a text fragment $w_1,...,w_n$ (a \emph{context}) and an NFH
anchor $a=(i,j)$ within it, we seek the head(s) of the anchor.

For \emph{Implicit} FH, the head can be any arbitrary expression. While our annotated corpus supports this (Section \ref{nfh-res-corpus}), in practice our modeling (Section \ref{sec:nfh-res}) as well as the annotation procedure favors selecting one out of 5 prominent categories or the \textsc{Other} category.

For \emph{Reference} FH, the head is selected from the text fragment. In principle a head can span multiple tokens (e.g. `unexpected thought' in (xii)).
This is also supported by our annotation procedure.
In practice, we take the syntactic head of the multi-token answer to be the single-token missing element, and defer the boundary resolution to future work.

In case multiple heads are possible for the same anchor (e.g. (viii,xiv) in Table 1), all should be recovered. Hence, the resolution task is a function from a (text, anchor) pair to a list of heads, where each head is either a single token in the text or an arbitrary expression.


\section{Numeric Fused-Head Identification}
\label{sec:nfh-iden}
The FH task is composed of two sub-tasks. In this section, we describe the first
\emp: identifying NFH \textit{anchors} in a sentence. We begin with a rule-based
method, based on the FH definition. We then proceed to a learning-based model, which achieves better results.

\paragraph{Test-set} We create a test set for assessing the identification methods by randomly
collecting 500 dialog fragments with numbers, and labeling each number as NFH or
not NFH.  We observe that \textbf{more than 41\% of the test-set numbers are FHs},
strengthening the motivation for dealing with the NFH phenomena.

\subsection{Rule-based Identification}
\label{sec:detect-num-fh}

FHs are defined as NPs in which the head is fused with a dependent element, resulting in an NP without a noun.\footnote{One exception are numbers which are part of names (`\emph{Appollo \textbf{11}'s your secret weapon?}') which we do not consider to be NFH.}
With access to an oracle constituency tree, NFHs can be easily identified by looking for such NPs. In practice, we resort to using automatically produced parse-trees.
\ignore{
Due to the lack of particular awareness of the fused-head phenomena and parser errors, the above rule makes relatively many mistakes. For example in the sentence: 

`The intimate details of \textbf{one hundred and sixty-three} confirmed kills.' the number in bold was tagged as a Noun.\footnote{Even though the PTB guidelines instruct to label some numbers as Nouns, this seems to be the case due to the fused-head phenomena and the wish to maintain on Nouns inside NPs}

We thus add heuristics on top of the main rule for achieving a more accurate result.
}

We parse the text using the Stanford constituency parser \cite{chen2014fast} and look for noun-phrases\footnote{Specifically, we consider phrases of type \textsc{np}, \textsc{qp}, \textsc{np-tmp}, \textsc{nx} and \textsc{sq}.} which contain a number but not a noun. This already produces reasonably accurate results, but we found that we can improve further by introducing 10 additional text-based patterns, which were customized based on a development set. These rules look for common cases that are often not captured by the parser.
For example, a conjunction pattern involving a number followed by `or', such as \textit{``eight or nine clubs''}\footnote{This phrase can be treated as a gapped coordination construction. For consistency, we treat it and similar cases as NFH, as discussed in Section \ref{sec:modeling-scope}. Another reading is that the entire phrase ``eight or nine'' refers to a single approximate quantity that modifies the noun ``clubs'' as a single unit. This relates to the problem of  disambiguating distributive-vs-joint reading of coordination, which we consider to be out of scope for the current work.}, where `eight' is an NFH which refers to `clubs'. 


Parsing errors result in false-positives. For example in ``You've had [\textbf{one} too many \textbf{cosmos}].'', the Stanford parser analyzes `one' as an NP, despite the head (`cosmos') appearing two tokens later. We cover many such cases by consulting with an additional parser. We use the \textsc{spaCy} dependency parser \cite{honnibal-johnson:2015:EMNLP} and filter out cases where the candidate anchor has a noun as its syntactic head or is connected to its parent via a \emph{nummod} label. We also filter cases where the number is followed or preceded by a currency symbol.

\paragraph{Evaluation}

We evaluate the rule-based identification on our test set, resulting in 97.4\% precision and 93.6\% recall.
The identification errors are almost exclusively a result of parsing mistakes in the underlying parsers. An example of a false-negative error is in the sentence: ``\emph{The lost
\textbf{six} belong in Thorn Valley.}'', where the dependency parser mistakenly labeled `belong' as a noun, resulting in a negative classification.
An example of a false-positive error is in the sentence: ``\emph{our God is the \textbf{one} true God}'' where the dependency parser labeled the head of \textbf{one} as `is'.

\subsection{Learning-based Identification}
\label{nfh-identification-ml}
We improve the NFH identification using machine learning. We create a large but noisy data set by considering all the numbers in the corpus and treating the NFH identified by the rule-based approach as positive (79,678 examples) and all other numbers as negative (104,329 examples).
We randomly split the dataset into train and development sets in a 90\%, 10\%
split.
Table \ref{tbl:det-fh-stats} reports the dataset size statistics.

We train a linear SVM classifier\footnote{{\tt sklearn} implementation \cite{pedregosa2011scikit} with default parameters.} with 4 features: (1) Concatenation of the anchor-span tokens; (2) Lower-cased tokens in a 3-token window surrounding the anchor span; (3) POS tags of tokens in a 3-token window surrounding the anchor span; and (4) POS-tag of the syntactic head of the anchor.
The features for the classifier require running a POS tagger and a dependency parser. These can be omitted with a small performance loss (see Table \ref{tbl:is-fh-res} for an ablation study on the dev set).

On the manually labeled test set, the full model achieves accuracies of 97.5\% precision and 95.6\% recall, \textbf{surpassing the rule-based approach}. 

\begin{table}[]
\begin{center}

\begin{tabular}{l|cc|c||c}
      & train 	& dev 	& test 	& all  	\\ \hline

pos	  & 71,821 	& 7865 	 & 206	& 79,884 \\
neg   & 93,785	& 10,536 & 294 & 104,623 \\ \hline \hline
all	  & 165,606 & 18,401 & 500 & 184,507
\end{tabular}
\end{center}
\caption{NFH Identification corpus summary. The train and dev splits are noisy and the test set are gold annotations.}
\label{tbl:det-fh-stats}
\end{table}

\subsection{NFH Statistics}
We use the rule-based positive examples of the dataset and report some statistics regarding the NFH phenomenon.
The most common anchor of the NFH-dataset with a very big gap is the token
`one'\footnote{Lower cased.} with 48,788 occurrences (61.0\% of the data), while
the second most commons is the token `two' with 6,263 occurrences (8.4\%).
There is a long tail in terms of the tokens occurrences, with 1,803 unique anchor tokens (2.2\% of the NFH-dataset).
Most of the anchors consist of a single token (97.4\%), 1.3\% contain 2 tokens and
the longest anchor consists of 8 tokens (`Fifteen million sixty one thousand and seventy six.').
The numbers tend to be written as words (86.7\%) and the rest are written as digits (13.3\%).

\begin{table}[]
\begin{center}

\begin{tabular}{l|ccc}

			& Precision		& Recall		& F1 \\ \hline
Deterministic (Test)				& 97.4         &  93.6  &  95.5 \\
Full-model (Test)				& \textbf{97.5}         &  \textbf{95.6}         &  \textbf{96.6} \\
\hline
Full-model (Dev)				& \textbf{96.8}         &  \textbf{97.5}         &  \textbf{97.1} \\
 - dep 				& 96.7         &  97.3         &  97.0 \\
 - pos 				& 96.4         &  97.0         &  96.7 \\
 - dep, pos			& 95.6         &  96.1         &  95.9 \\ 
 
\end{tabular}
\end{center}
\caption{NFH Identification results}
\label{tbl:is-fh-res}
\end{table}

\subsection{NFH Identification Dataset}
The underlying corpus contains 184,507 examples (2,803,009 tokens), of which 500
examples are gold-labeled and the rest are noisy. In the gold test set, 41.2\%
of the numbers are NFHs. The estimated quality of the corpus---based on the
manual test-set annotation---is 96.6\% F1 score. The corpus and the NFH identification models are available at \url{github.com/yanaiela/num_fh}.

\section{NFH Resolution Dataset}
\label{nfh-res-corpus}

Having the ability to identify NFH cases with high accuracy, we turn to the more
challenging task of NFH resolution. The first step is creating a gold annotated dataset.

\subsection{Corpus Candidates}

Using the identification methods---which achieve satisfying results---we identify a total of 79,884 NFH cases in the IMDB corpus. We find that a large number of the cases follow a small set of patterns and are easy to resolve deterministically: four deterministic
patterns account for 28\% of the NFH cases. The remaining cases are harder. We
randomly chose a 10,000 subset of the harder cases for manual annotation via crowd-sourcing.
We only annotate cases where the rule-based and learning-based identification
methods agree.

\begin{table*}[t]
\begin{center}
\footnotesize{
\begin{tabular}{l|l|c|c}
Pattern           & Example	& Head	& Frequency (\%)                            \\ \hline
no \textbf{one}         & 
\includegraphics[scale=0.7]{trees/no-one.tex}
& \textsc{People} & 6.8 \\

you \textbf{two}        & 
\includegraphics[scale=0.7]{trees/you-two.tex}
& \textsc{People} & 2.3 \\

\textbf{NUM} \underline{of} \textit{NP} & 
\includegraphics[scale=0.7]{trees/one-of-tree.tex} & \textit{dreams} & 15.8       \\

\textit{X} \underline{`be' the} \textbf{one}        & 
\includegraphics[scale=0.7]{trees/be-the-one-tree.tex} &  \textit{Theresa} & 2.9 \\

\end{tabular}
}
\end{center}

\caption{Example of NFH whose heads can be resolved deterministically. The first two patterns are the easiest to resolve. These just have to match as is and their head is the PEOPLE class. The last two patterns depends on a dependency parser and can be resolved by following arcs on the parse tree.}
\label{tbl:easy-fh}
\end{table*}

\paragraph{Deterministic Cases}
The four deterministic patterns along with their coverage are detailed in Table \ref{tbl:easy-fh}. The first two are straightforward string matches for the
patterns \emph{no one} and \emph{you two}, which we find to almost exclusively
resolve to \textsc{People}. The other two are dependency-based patterns for
partitive (\emph{four [children] of the children}) and copular (\emph{John is
the one [John]}) constructions.
We collected a total of 22,425 such cases. While we believe these cases need to be
handled by any NFH resolution system, we do not think systems should be evaluated
on them. Therefore, we provide these cases as a separate dataset.

\subsection{Annotation via Crowdsourcing} 
\label{dataset-creation}

The FH phenomenon is relatively common and can be understood easily by non-experts, making the task suitable for crowd-sourcing. 

\paragraph{The Annotation Task}
For every NFH anchor, the annotator should decide if it is a \emph{Reference FH}
or an \emph{Implicit FH}. For \emph{Reference}, they should mark the relevant textual span. For \emph{Implicit}, they should specify the implicit head from a closed list. 
In cases where the missing head belongs to the implicit list, but also appears as a span in the sentence (reference), the annotators are instructed to treat it as a reference.
To encourage consistency, we run an initial annotation in which we identified
common implicit cases: \textsc{Year} (a calendar year, example (ix) in Table 1),
\textsc{Age} (example (x)), \textsc{Currency} (example (xi). While the source of
the text suggests US dollars, we do not commit to a specific currency),
\textsc{Person/People} (example (vi)) and \textsc{Time} (a
daily hour, example (iii)).
The annotators are then instructed to either choose from these five categories; to
choose \textsc{Other} and provide free-form text; or to choose \textsc{Unknown} in
case the intended head cannot be reliably deduced based on the given
text.\footnote{This
happens, for example, when the resolution depends on another modality.
For example, in our setup using dialogues from movies and TV-series, the speaker could refer
to something from the video which isn't explicitly mentioned in the text, such
as in ``\emph{Hit the deck, Pig Dog, and give me \textbf{37}!}''.}
For the \emph{Reference} cases, the annotators can mark any contiguous span in
the text. We then simplify their annotations and consider only the syntactic
head of their marked span.\footnote{We do provide the entire span annotation as
well, to facilitate future work on boundary detection.} This could be done automatically in most cases, and
was done manually in the few remaining cases. The annotator must choose a single
span. In case the answer comprises of several spans as in examples (viii, xiv),
we rely on it to surface as a disagreement between the annotators, which we then pass to further resolution by expert annotators.

\paragraph{The Annotation Procedure}
We collected annotations using Amazon Mechanical Turk (AMT).\footnote{To
maximize the annotation quality, we restricted the turkers with the following
requirements: Complete over 5K acceptable HITs, Over 95\% of their overall HITs
being accepted and Completing a qualification for the task.}
In every task (HIT in AMT jargon) a sentence with the FH \textit{anchor} was
presented (\textit{target sentence}). Each \textit{target sentence} was
presented with maximum two dialog turns before and one dialog turn after it. This was the sole context that was shown for avoiding to exhaust the AMT workers (turkers) with long texts and in the vast majority of the examined examples, the answer appeared in that scope.

\begin{figure}[h]
\centering
\includegraphics[scale=0.14]{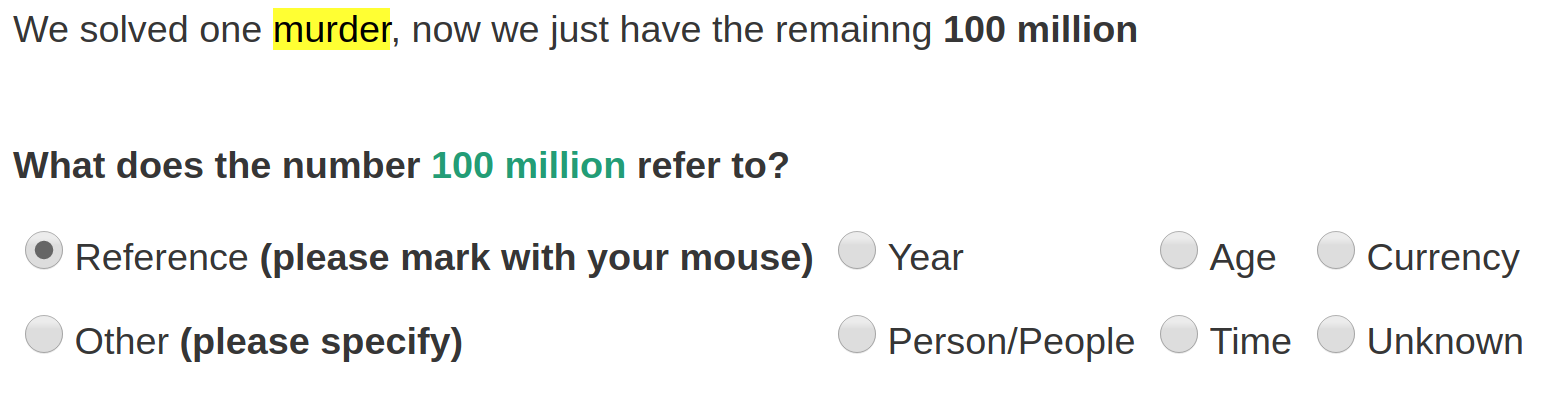}
\caption{Crowdsourcing Task Interface on AMT}
\label{fig:task}
\end{figure}

Every HIT contained a single NFH example. In cases of more than one NFH per sentence, it was split into 2 different HITs.
The annotators were presented with the question: ``What does the number [ANCHOR]
refer to?'' where [ANCHOR] was replaced with the actual number span, and were
asked to choose from 8 possible answers: \textsc{Reference}, \textsc{Year},
\textsc{Age}, \textsc{Currency}, \textsc{Person/People}, \textsc{Time},
\textsc{Other} and \textsc{Unknown} (See Figure \ref{fig:task} for a HIT example).
Choosing the \textsc{Reference} category requires marking a span in the
text, corresponding to the referred element (the missing head). The turkers were
instructed to prefer this category over the others if possible. Therefore, in example (xiv) the \textit{Reference} answers were favoured over the \textsc{People} answer. 
Choosing the \textsc{Other} category required entering free-form text.

Post-annotation, we unify the \emph{Other} and \emph{Unknown} cases into a single
\textsc{Other} category.

\begin{figure}[h]
    \centering
    \includegraphics[scale=0.6]{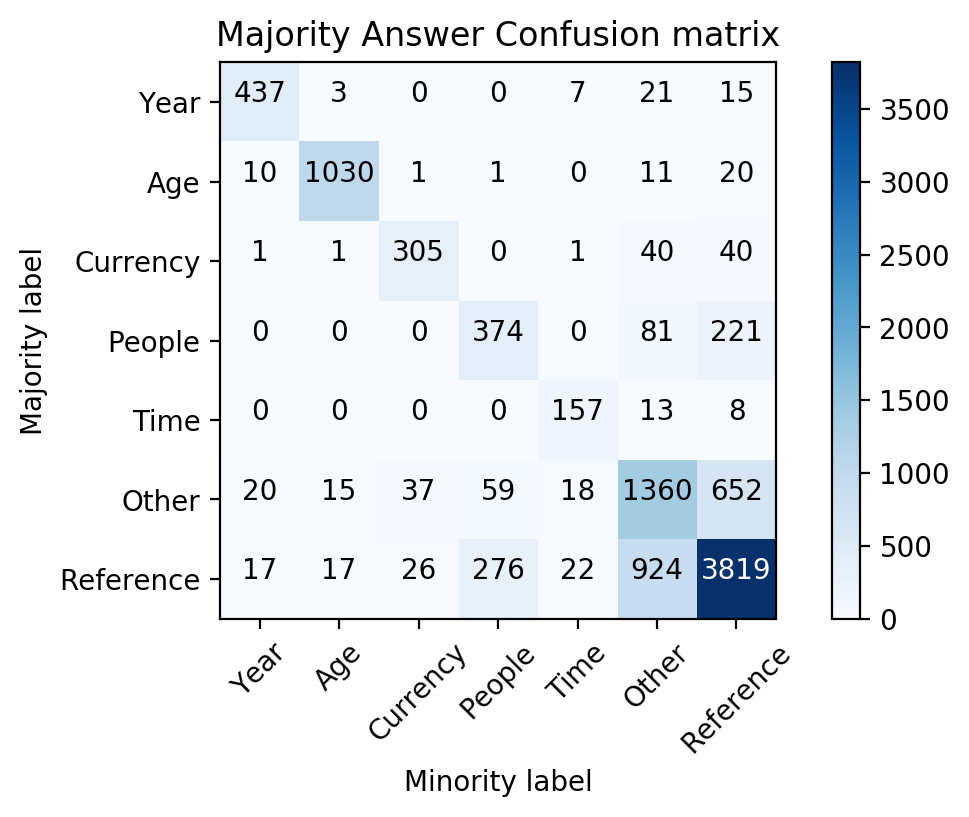}
     \caption{Confusion Matrix of the majority annotators 
     on categorical decision}
     \label{fig:ans-agg} 
\end{figure}

Each example was labeled by three annotators.
On the categorical decision (just the 1-of-7 choice, without considering the spans selected for the \textsc{Reference} text and combining the \textsc{Other} and \textsc{Unknown} categories), 73.1\% of the cases had a perfect agreement (3/3), 25.6\% had a majority agreement (2/3), and 1.3\% had a complete disagreement.
The Fleiss kappa agreement \cite{fleiss1971measuring} is $k=0.73$,
a substantial agreement score.
The high agreement score suggests that the annotators tend to agree on the
answer for most cases.
Figure \ref{fig:ans-agg} shows
the confusion matrix for the 1-of-7 task, excluding the cases of complete
disagreement. 
The more difficult cases involve the \textsc{Reference} class, which is often
confused with \textsc{People} and \textsc{Other}.

\subsection{Final Labeling Decisions}
Post-annotation, we ignore the free text entry for \textsc{Other} and
unify \textsc{Other} and \textsc{Unknown} into a single category. However, our
data collection process (and the corpus we distribute) contain this information,
allowing for more complex task definitions in future work.

The disagreement cases surface genuinely hard cases, such as the ones below:
\begin{exe}
\ex\label{ex:hard} Mexicans have \textbf{fifteen}, Jews have thirteen, rich girls have sweet sixteen...
\ex\label{ex:no-lemma-agg} All her \textit{communications} are to Minnesota \textit{numbers}. There's not \textbf{one} from California.
\ex\label{ex:fh-hard} And I got to see \textit{Irish}. I think he might be the \textbf{one} that got away, or the one that got put-a-way.
\end{exe}

The majority of the partial category agreement cases (1576) are of \textsc{Reference} vs.\ \textsc{Other}/\textsc{Unknown}, which are indeed quite challenging (e.g. Example \ref{ex:fh-hard} where two out of three turkers selected the \textsc{Reference} answer and marked \textit{Irish} as the head, and the third turker selected the Person/People label, which is also true, but less meaningful in our perspective).

The final labeling decision was carried out in two phases.
First, a categorical labeling was done using the majority label, while the 115
examples with disagreement (e.g. Example \ref{ex:hard} which was tagged as
\textsc{Year}, \textsc{Reference} (`birthday' which appeared in the context) and
\textsc{Other} (free text:`special birthday') were annotated manually by experts.

The second stage dealt with the \textsc{Reference} labels (5718 cases).
We associate each annotated span with the lemma of its syntactic head,
and consider answers as equivalent if they share the same lemma
string.
This results in 5101 full-agreement cases on the lemma level.
The remaining 617 disagreement cases
(e.g. Example (\ref{ex:no-lemma-agg})) were passed to further annotation by the expert annotators.
During the manual annotation we allow also for multiple heads for a single
anchor (e.g. for (viii,xiv) in Table \ref{tbl:examples}).

An interesting case in \emph{Reference} FH are constructions in which the
referenced head is not unique. Consider example (viii) in Table \ref{tbl:examples}:
the word `one' refers to either \emph{men} or \emph{busses}.
Another example of such case is example (xiv) in Table \ref{tbl:examples} where the word `two' refers both to \emph{fussy old maid} and to \emph{flashy young man}.
Notice that the two cases have different interpretations: the referenced heads in (viii) have an \textit{or} relation between them whereas the relation in (xiv) is \textit{and}. 

\subsection{NFH Statistics}

\begin{figure}[]
    \centering
    \includegraphics[scale=0.045]{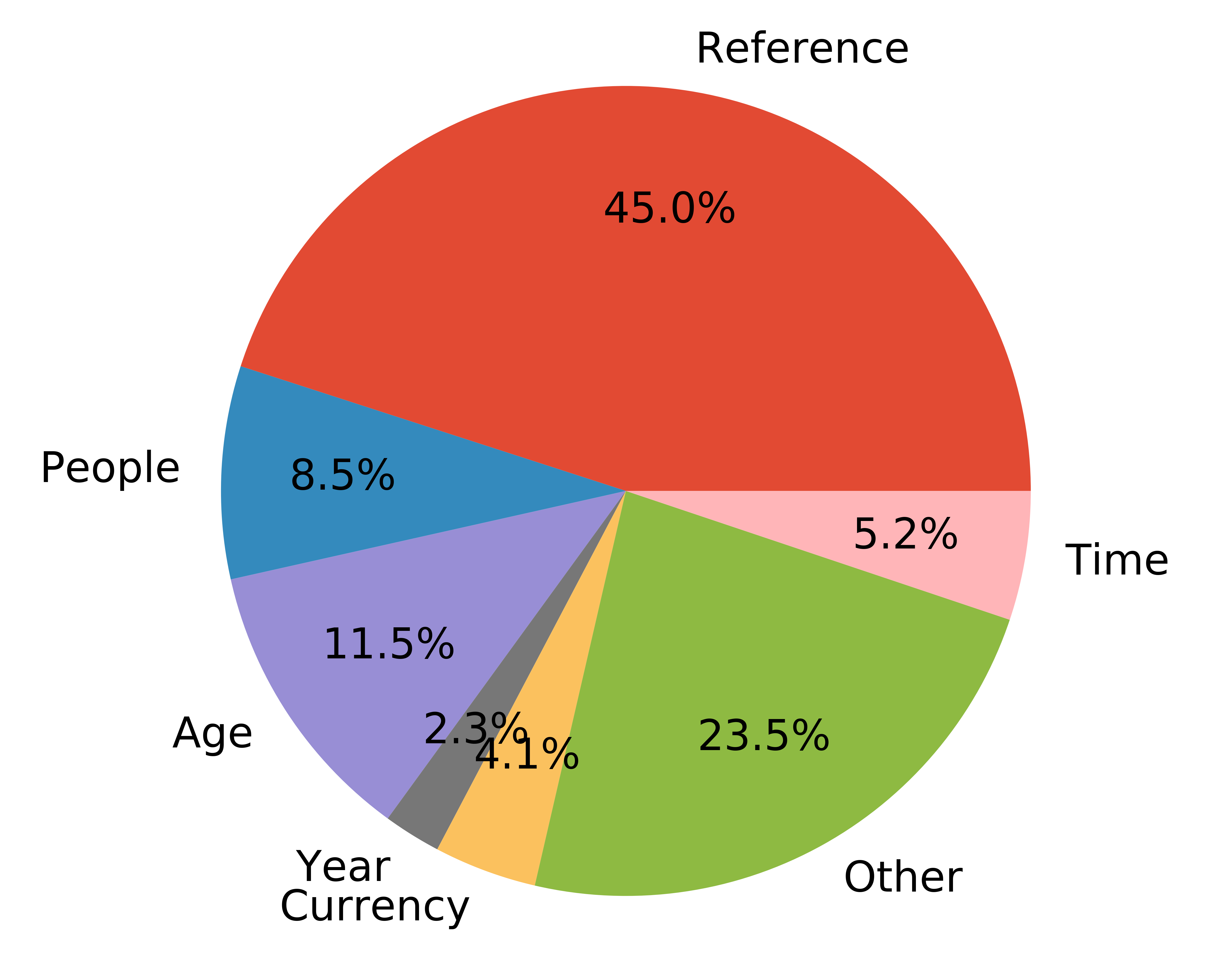}
     \caption{Distribution of NFH types in the NFH-Resolution dataset.}
     \label{fig:cls-dist}
\end{figure}

\paragraph{General}
We collected a total of 9,412 annotated NFH. The most common class is \textsc{Reference} (45.0\% of the dataset). The second common class is
\textsc{Other} (23.5\%), which is the union of original \textsc{Other} class, in which
turkers had to write the missing head, and the \textsc{Unknown} class, in which no clear answer could be identified in the text. The majority of this joined class is from the \textsc{Unknown} label (68.3\%). 
The rest of the 5 closed-class categories account for the other 31.5\% of the cases. A full
breakdown is given in Figure \ref{fig:cls-dist}.
The \textit{anchor} tokens in the dataset mainly consist of the token `one' (49.0\% of the dataset), with the tokens `two' and `three' being the second and third most common. 377 (3.9\%) of the \textit{anchors} are singletons, which appear only once.

\paragraph{Reference Cases}
The dataset consists of a total of 4,237 \textsc{Reference} cases. The vast majority of them (3,938 cases) were labeled with a single referred element, 238 with two reference-heads and 16 with three or more.

In most of the cases, the reference span can be found near the anchor span. In 2,019 of the cases, the reference is in the same sentence with the anchor, in 1,747 it appears in a previous/following sentence.
Furthermore, in most cases (82.7\%), the reference span appears before the anchor and only in 5.1\% of the cases it appears after it. An example of such a case is presented in Example (xiv) in Table \ref{tbl:examples}. In the rest of the cases, references appear both before and after the anchor.

\subsection{NFH Resolution Dataset}
The final NFH Resolution dataset consists of 900,777 tokens containing 9,412 instances of gold labeled resolved NFH. The resolution was done by 3 mechanical turk annotators per task, with a high agreement score ($k=0.73$)\footnote{The \textit{Reference} cases were treated as a single class for computing the agreement score.}
The \textsc{Reference} cases are annotated with at least one referring item. The
\textsc{Other} class unifies several other categories (None and some other
scarce \textit{Implicit} classes), but we maintain the original turkers answers
to allow future work to apply more fine-grained solutions for these cases.

\section{Where's my head? Resolution Model}
\label{sec:nfh-res}

We consider the following resolution task: given a numeric anchor and its
surrounding context, we need to assign it a single head. The head can be either
a token from the text (for \emph{Reference FH}) or one-of-six categories (the 5 most common categories and \textsc{Other}) for \emph{Implicit FH}.\footnote{This is a somewhat simplified version of the full task defined in Section \ref{sec:nfh-modeling}. In
particular, we do not require to specify the head in case of \textsc{Other},
and we require a single head rather than a list of heads. Nonetheless, we find
this variant to be both useful and challenging in practice. For the few multiple-head cases, we consider each of the items in the gold list to be correct, and defer a fuller treatment for future work.}

This combines two different kinds of tasks.
The \textsc{Reference} case requires selecting the most adequate token over the text, suggesting a similar formulation to coreference resolution
\cite{ng2010supervised,lee2018higher} and  implicit arguments identification \cite{gerber2012semantic,mooretal:13}. 
The implicit case requires selection from a closed list, a similar formulation to word-tagging-in-context tasks, where the word (in our case span) to be tagged is the anchor. A further complication is the need to weigh the different decisions (\textit{Implicit} vs \textit{Reference}) against each other.
Our solution is closely modeled after the state-of-the-art coreference resolution system of \citet{lee2017end}.\footnote{Newer systems such as \citet{lee2018higher}, \citet{zhang2018neural} show improvements on the coreference task, but using components which focus on the clustering aspect of coreference, which are irrelevant for the NFH task.}
However, the coreference-centric architecture had to be adapted to the particularities of the NFH task.
Specifically, (a) the NFH resolution does not involve cluster assignments, and (b) it requires handling the \emph{Implicit} cases in addition to the \emph{Reference} ones.

The proposed model combines both decisions, a combination which resembles the copy-mechanisms in neural MT \cite{gu2016incorporating} and the Pointer Sentinel Mixture Model in neural LM \cite{merity2016pointer}.
As we only consider referring mentions as single tokens, we discarded the original models' features which handled the multi-span representation (e.g. the Attention mechanism).
Furthermore, as the Resolution task already receives a numeric anchor, it is
redundant to calculate a mention score. In preliminary experiments we did try to
add an antecedent score, with no resulting improvement. Our major adaptations to
the \citet{lee2017end} model, described below, are the removal of the redundant components, and the
addition of an embedding matrix for representing the \textit{Implicit} classes.

\subsection{Architecture}

Given an anchor, our model assigns a score to each possible anchor-head pair and
picks the one with the highest score. The head can be either a token from the text (for the \textit{Reference} case) or one-of-six category labels (for the \textit{Implicit} case).
We represent the anchor, each of the text tokens and each category label as vectors.

Each of the implicit classes $c_1,...,c_6$ is represented as an embedding vector
$\mathbf{c_i}$, which is randomly initialized and trained with the system.

To represent the sentence tokens ($\mathbf{t_i}$), we first represent each token as a concatenation of the token embedding and the
last state of a character LSTM \cite{hochreiter1997long}:
\[\boldsymbol{x_i}=[\mathbf{e_i};LSTM(\mathbf{e_{i_{c1:ct}}})]\]
where $\mathbf{e_i}$ is the $i$th token embedding and $\mathbf{e_{i_{c_j}}}$ is
the $j$th character of the $i$th token.
These representations are then fed into a text-level biLSTM resulting in the
contextualized token representations $\mathbf{t_i}$:

\[\mathbf{t_i} = BILSTM(\mathbf{x_{1:n}},i)\]

Finally, the \emph{anchor}, which may span several tokens, is represented as the average
over its contextualized tokens.
\[\boldsymbol{a} = \frac{1}{j - i + 1} \sum_{k=i}^{j} \boldsymbol{t_k} \]

We predict a score $s(h,a)$ for every possible head-anchor pair, where
$h\in \{c_1,...,c_6,t_1,...,t_n\}$ and $\mathbf{h_i}$ is the corresponding
vector.
The pair is represented as a concatenation of the head, the anchor and their
element-wise multiplication, and scored with a multi-layer perceptron:
\[s(h,a)= MLP([\textbf{h};\textbf{a};\textbf{h} \odot \textbf{a}])\]

We normalize all of the scores using softmax, and train to minimize the cross-entropy loss.

\paragraph{Pre-trained LM}
To take advantage of the recent success in pre-trained language models \cite{peters2018deep,devlin2018bert} we also make use of \texttt{ELMo} contextualized embeddings instead of the embedding matrix and the character LSTM concatentation.

\subsection{Training Details}
The character embedding size is 30 and their LSTM dimension is 10. We use Google's pre-trained 300-dimension w2v embeddings \cite{mikolov2013distributed} and fix the embeddings so they don't change during training.
The text-level LSTM dimension is 50. The \textit{Implicit} embedding size is the same as the BiLSTM output, 100 units.
The MLP has a single hidden layer of size 150 and uses $tanh$ as the
non-linear function. We use dropout of 0.2 on all hidden layers, internal
representation and tokens representation. We train using the Adam optimizer
\cite{kingma2015j} and a learning rate of 0.001 with early stopping, based on the development set. We shuffle the training data before every epoch. The annotation allows more than one referent answer per anchor, in such case, we take the closest one to the anchor as the answer for training, and allow either one when evaluating.
The experiments using \texttt{ELMo}, replaced the pre-trained word embeddings and character LSTM. It uses the default parameters in the AllenNLP framework \cite{Gardner2017AllenNLP}, with 0.5 dropout on the network, without gradients update on the contextualized representation.

\subsection{Experiments and Results}
\label{sec:experiments-results}
\begin{table}[]
\centering
\begin{tabular}{l|c|c}
Model     & Reference & Implicit \\ \hline

Oracle (Reference) & 70.4 & - \\
{+} Elmo		   & 81.2   & - \\
Oracle (Implicit)  & - & 82.8 \\
{+} Elmo           & - & 90.6 \\ \hline \hline
Model (full)  	   &  61.4 & 69.2 \\
{+} Elmo           & 73.0 & 80.7
\end{tabular}
\caption{NFH Resolution accuracies for the \textit{Reference} and
\textit{Implicit} cases on the development set. \textbf{Oracle (Reference)} and \textbf{Oracle
(Implicit)} assume an oracle for the implicit vs. reference decisions. \textbf{Model (full)} is our final model.}
\label{tbl:baseline}
\end{table}

\paragraph{Dataset Splits}
We split the dataset into train/development/test, containing 7,447, 1,000 and 1,000
examples, respectively. There is no overlap of movies/TV-shows between the different splits.

\paragraph{Metrics}
We measure the model performance of the NFH head detection using accuracy. For
every example, we measure if the model successfully predicted the correct label or not. We report two additional measurements: binary classification accuracy
between the \textit{Reference} and \textit{Implicit} cases and a multiclass
classification accuracy score, which measures the class-identification accuracy
while treating all \textsc{Reference} selections as a single decision,
regardless of the chosen token.

\paragraph{Results}
We find that 91.8\% of the \textit{Reference} cases are nouns.
To provide a simple baseline for the task, we report accuracies solely on the \textit{Reference} examples (ignoring the \textit{Implicit} ones) when choosing one of the surrounding nouns.
Choosing the first noun in the text, the last one or the closest one to the anchor leads to scores of 19.1\%, 20.3\% and 39.2\%

We conduct two more experiments to test our model on the different FH kinds:
\textit{Reference} and \textit{Implicit}. In these experiments we assume an oracle that tells us the head type (\textit{Implicit} or \textit{Reference}) and restricts the candidate set for the correct kind during both training and testing.
Table \ref{tbl:baseline} summarizes the results for the oracle experiments as well as for the full model.

\begin{table}[]
\centering
\begin{tabular}{l|c|c}
Model     & Development & Test \\ \hline

Base & 65.6 & 60.8 \\
{+} Elmo		   & \textbf{77.2}   & \textbf{74.0} \\
\end{tabular}
\caption{NFH Resolution accuracies on the development and test sets.}
\label{tbl:nfh-resuolution-res}
\end{table}

The final models accuracies are summarized in Table \ref{tbl:nfh-resuolution-res}. 
The complete model trained on the entire training data achieves 65.6\% accuracy on the development set and 60.8\% accuracy on the test set.
The model with \texttt{ELMo} embeddings \cite{peters2018deep} adds a significant boost in performance and achieves 77.2\% and 74.0\% accuracy on the development and test sets respectively.

The development-set binary separation with \texttt{ELMo} embeddings is 86.1\% accuracy and categorical separation is 81.9\%. 
This substantially outperforms all baselines, but still lags behind the oracle
experiments (\textit{Reference}-only and \textit{Implicit}-only).

As the oracle experiments perform better on the individual \textit{Reference} and \textit{Implicit} classes, we experimented with adding an additional  objective to the model which tries to predict the oracle decision (implicit vs. reference). This objective was realized as an additional loss term. However, this experiment didn't yield any performance improvement.

We also experimented with linear models, with features based on previous work
which dealt with antecedent determination \cite{ng2005machine,liu2016exploring}
such as POS tags and dependency labels of the candidate head, if the head is the
closest noun to the anchor etc. We also added some specific features that dealt
with the Implicit category, for example binarization of the anchor based on its
magnitude (e.g. <1,<10,<1600,<2100), if there was another currency mention in
the text, etc. None of these attempts surpassed the 28\% accuracy on the
development set. For more details on these experiments, see Appendix \ref{sec:appendix-linear}.

\subsection{Analysis}
The base model's results are relatively low, but gain a substantial improvement by adding contextualized embeddings. We perform an error analysis on the \texttt{ELMo} version which highlights the challenges of the task.

\begin{figure}[h]
    \centering
    \includegraphics[scale=0.6]{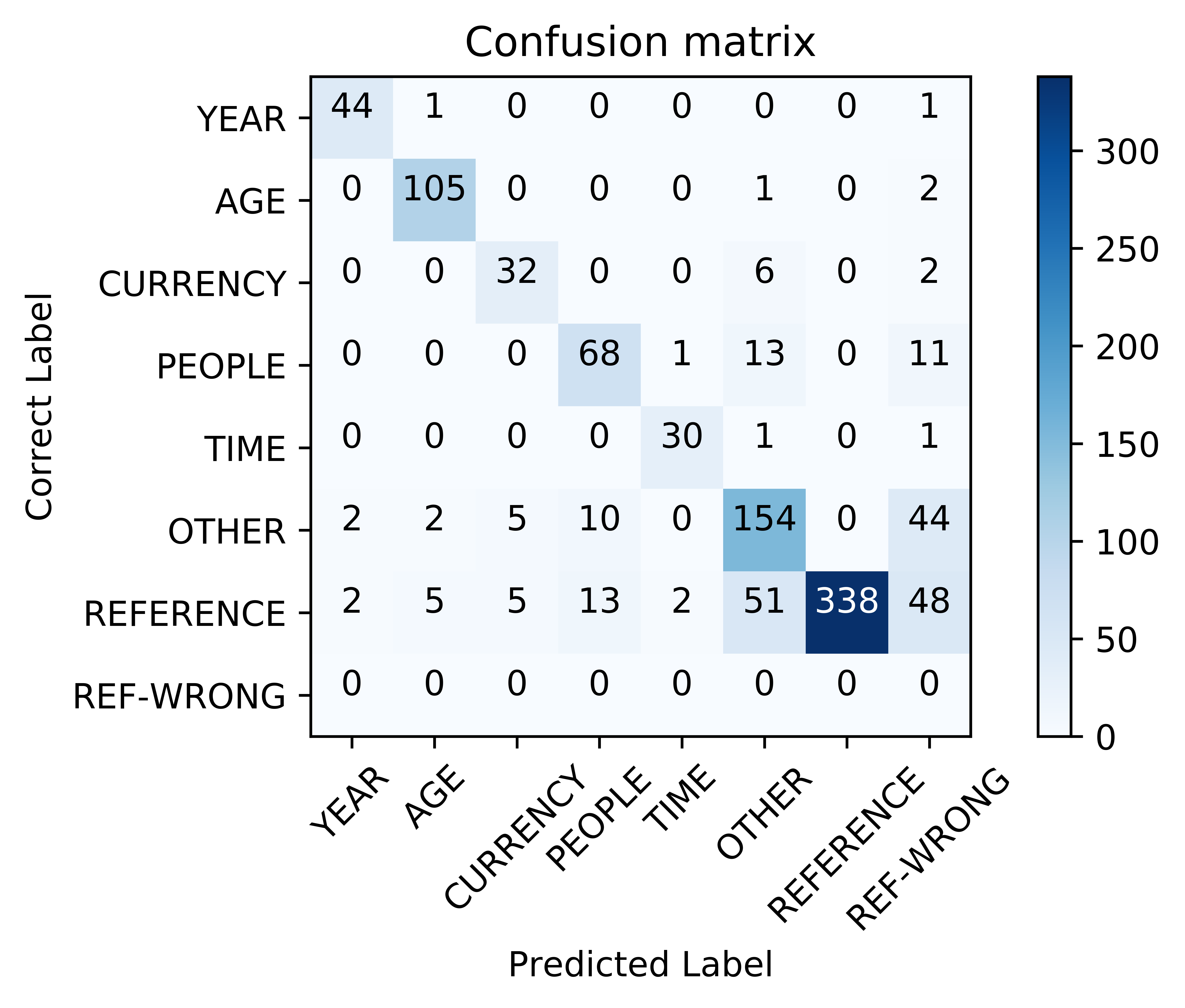}
     \caption{Confusion Matrix of the model. Each row/column corresponds to a gold/predicted label respectively. The last one (REF-WRONG), is used for indicating an erroneous choice of a \textit{Reference} head.}
     \label{fig:res-conf} 
\end{figure}

\begin{table*}[t]
\begin{center}
\resizebox{\textwidth}{!}{
\begin{tabular}{l|l|l|l}

 & Text & Predicted & Truth \\ \hline
1 & \begin{tabular}[c]{@{}l@{}}\textbf{Dreadwing:} This will be my gift to the Dragon Flyz, my farewell \colorbox{choice_medium}{\textit{\underline{gift}}}.\\\textbf{\color{blue}{One}} that will keep giving and giving and giving.\end{tabular} & \textsc{People} & \textit{gift} \\ \hline

2 & \begin{tabular}[c]{@{}l@{}}\textbf{David Rossi:} How long? \\
 \textbf{Harrison Scott:} A \colorbox{choice_medium}{year}. Maybe \textbf{\color{blue}five}. It's hard to keep track without a watch.\end{tabular} & \textsc{Age} & \textsc{Year} \\ \hline

3 & \begin{tabular}[c]{@{}l@{}}\textbf{Henry Fitzroy:} a hundred \textit{\underline{dollars}}, that's all it takes for you to risk your \colorbox{choice_medium}{life}? \\
\textbf{Vicki Nelson:} Actually, it was more around \textbf{\color{blue}{98}}...
 
 \end{tabular} & \textsc{Other} & \textit{dollar} \\ \hline \hline

4 & \begin{tabular}[c]{@{}l@{}} \textbf{Evelyn Pons:} He might be my legal \colorbox{choice_medium}{\textit{\underline{dad}}}, too! \\
\textbf{Paula Novoa Pazos:} No, because we're not \colorbox{choice_strong}{sisters}, but you can look for another one.\\
\textbf{Evelyn Pons:} How did you look for \textbf{\color{blue}{one}}? \end{tabular} & \textit{sisters} & \textit{dad} \\ \hline




5 & \begin{tabular}[c]{@{}l@{}} \textbf{L'oncle Irvin:} A \colorbox{choice_medium}{\textit{\underline{soul}}}.\\
\textbf{Krank:} Because you believe you have \textbf{\color{blue}{one}}? You don't even have a \colorbox{choice_strong}{body}.
\end{tabular} & \textit{body} & \textit{soul} \\ \hline

 
6 & \begin{tabular}[c]{@{}l@{}}\textbf{Jenny:} Head in the clouds, that \textbf{\color{blue}{one}}. I don't know why you're so sweet on him.
 \end{tabular} & \textsc{Other} & \textsc{People} \\ \hline
 
7 & \begin{tabular}[c]{@{}l@{}} \textbf{Officer Mike Laskey:} I can't do that. \\
\textbf{Joss Carter:} Do you really wanna test me? 'Cause I've got a shiny new \textbf{\color{blue}{1911}} [...] \\ \hline
\end{tabular} & \textsc{Year} & \textsc{Other} \\ \hline
 
\end{tabular}
}
\end{center}

\caption{Erroneous example predictions from the development data. Each row represents an example from the data. The redder the words, the higher their scores. The two last columns contain the model prediction and the gold label. Uppercase means the label is from the \textsc{Implicit} classes, otherwise it's a \textsc{Reference} in lowercase.}
\label{tbl:err-ex}
\end{table*}

Figure \ref{fig:res-conf} shows the confusion matrix of our model and Table \ref{tbl:err-ex} lists some errors from the development set.

\paragraph{Pattern-Resolvable Error-cases}
The first three examples in Table \ref{tbl:err-ex} demonstrate error cases that can be solved based on text-internal cues and ``complex-pattern-matching'' techniques. These can likely be improved with a larger training set or improved neural models.

The errors in Examples 1 and 2 might have caused by a multi-sentence patterns. A possible reason for the errors is due to the lack of that pattern in the training data. Another explanation could be due to a magnitude bias, where in Example 1, \textbf{One} in the beginning of a sentence usually refer to \textsc{People}, whereas in Example 2, \textbf{Five} is more likely to refer to an \textsc{Age}.

In Example 3, the model has to consider several cues from the text, such as the phrase ``\textit{a hundred dollars}'' which contains the actual head and is of a similar magnitude to the anchor. In addition, the phrase: ``\textit{it was more around}'' gives a strong hint on a previous reference.

\paragraph{Inference/Common Sense Errors}
Another category of errors is those that are less likely to be resolved with pattern-based techniques and more data. These require common sense and/or more sophisticated inferences to get right, and will likely require a more sophisticated family of models to solve.

In Example 4, \textbf{one} refers to \textit{dad}, but the model chose \textit{sisters}. These are the only nouns in this example, and with the lack of any obvious pattern, a model needs to understand the semantics of the text to identify the missing head correctly.

Example 5 also requires to understand the semantics of the text, and some understanding of its discourse dynamic; where a conversation between the two speakers takes place, with a reply of \textit{Krank} to \textit{L'oncle Irvin}, that the model missed.

In Example 6, the model has difficulty to collect the cues in the text that refer to an unmentioned person, and therefore the answer is \textsc{People}, but the model predicts \textsc{Other}.

Finally, in Example 7 we observe an interesting case of overfitting, which is likely to originate from the word-character encoding. As the anchor - \textbf{1991} is a four-digit number, which are usually used to describe \textsc{Year}s, its representation gets a strong signal for this label, even though the few words which precede it (\textit{a shiny new}) are not likely to describe a \textsc{Year} label.

\section{Related Work}
\label{sec:related}
The FH problem was not directly studied in the NLP literature. However, several works dealt with overlapping components of this problem.

\paragraph{Sense Anaphora}
The first, and most related is the line of work by \citet{gardiner2003identifying,ng2005machine,recasens2016sense} which dealt with sense anaphoric pronouns (``Am I a \emph{suspect}? - you act like \textbf{one}'', c.f. Example \ref{ex4}).
\emph{Sense anaphora}, sometimes also referred to as \emph{identity of sense anaphora}, are expressions that inherit the sense from their antecedent but do not denote the same referent (as opposed to coreference). 
The sense anaphora phenomena cover also numerals, and significantly overlap with many of our NFH cases. However, it does not cover the \emph{Implicit} NFH cases, and also does not cover cases where the target is part of a co-referring expression (``I met \emph{Alice and Bob}. The \textbf{two} seem to get along well.'').

In terms of computational modeling, the sense anaphora task is traditionally split into two subtasks: (i) identifying anaphoric targets and disambiguating their sense; and (ii) resolving the target to an antecedent.
\citet{gardiner2003identifying} and \citet{ng2005machine} perform both tasks, but restrict themselves to \emph{one anaphora} cases and their noun-phrase antecedents. 
\citet{recasens2016sense} on the other hand addressed a wider variety of sense anaphors (e.g. \textit{one}, \textit{all}, \textit{another}, \textit{few}, \textit{most}, etc. a total of 15 different senses, including \emph{numerals}). \citet{recasens2016sense} annotated a corpus of third of the English OntoNotes \cite{weischedel2011ontonotes} with sense anaphoric pronouns and their antecedents.
Based on this dataset, they introduce a system for distinguishing anaphoric from non-anaphoric usages. However, they do not attempt to resolve any target to its antecedent.
The non-anaphoric examples in their work combines both our \textit{Implicit} class, as well as other non-anaphoric examples indistinguishably, and therefore are not relevant for our work.

In the current work, we restrict ourselves to numbers and so cover only part of the sense-anaphora cases handled in \citet{recasens2016sense}. However, in the categories we do cover, we do not limit ourselves to anaphoric cases (e.g. Ex. \ref{ex3}, \ref{ex4}) but include  also non-anaphoric cases that occur in FH constructions (e.g. Ex. \ref{ex1}, \ref{ex2}) and are interesting on their own right.
Furthermore, our models not only identify the anaphoric cases but also attempt to resolve them to their antecedent.

\paragraph{Zero Reference}
In \emph{zero reference}, the argument of a predicate is missing, but it can be easily understood from context \cite{hangyo2013japanese}.
 For example, in the sentence: \textit{``There are two roads to eternity, a straight and narrow \emp, and a broad and crooked \emp''} have a zero-anaphoric relationship to ``two roads to eternity'' \cite{iida2006exploiting}.
This phenomenon is usually discussed as the context of \emph{zero pronouns}, where a pronoun is what's missing. It occur mainly in pro-drop languages such as Japanese, Chinese and Italian, but has also observed in English, mainly in conversational interactions \cite{oh2005english}.
 Some, but not all, zero-anaphora cases result in FH or NFH instances.
Similarly to FH, the omitted element can appear in the text, similar to our
\emph{Reference} definition (zero \textbf{endophora}), or outside of it,
similar to our \emph{Implicit} definition (zero \textbf{exophora}).
Identification and resolution of this, has attracted a lot of interest mainly in
Japanese \cite{nomoto1993resolving,hangyo2013japanese,iida2016intra} and Chinese \cite{chen2016chinese,P18-1053,yin2018zero}, but also in other languages \cite{ferrandez2000computational,yeh2001empirical,han2004korean,kong2010tree,mihuailua2010or,kopec2014zero}.
However, most of these works considered only the zero endophora phenomenon in
their studies, and even those who did consider zero exophora
\cite{hangyo2013japanese}, only considered the author/reader mentions, e.g.
\textit{``liking pasta ($\phi$) eats ($\phi$) every day''} (translated from
Japanese). In this study, we consider a wider set of possibilities. Furthermore,
to the best of our knowledge, we are the first to tackle (a subset-of) zero anaphora in English.

\ignore{
\paragraph{Implicit arguments}
For example: Automatically Identifying Implicit Arguments to
Improve Argument Linking and Coherence Modeling, http://www.aclweb.org/anthology/S13-1043.pdf. seems very relevant}

\paragraph{Coreference}
The coreference task is to find within a document (or multiple documents) all
the corefering spans which form cluster(s) of the same mention (which are the
anaphoric cases as described above). The FHs resolution task, apart from the
non-anaphoric cases, is to find the correct anaphora reference of the target
span. The span identification component of our task overlaps with the coreference one (see \citet{ng2010supervised} for a thorough summary on the Noun Phrase coreference resolution and \cite{sukthanker2018anaphora} for a comparison between coreference and anaphora). Although the span search resemblance, the key conceptual distinctions is that FH allow the anaphoric span to be non co-referring


Recent work on coreference resolution \cite{lee2017end} propose an end-to-end neural architecture which results in a state-of-the-art performance. 
The work of \cite{peters2018deep,lee2018higher,zhang2018neural} further improve on their the scores with pre-training, refining span representation and using biaffine attention model for mention detection and clustering.
While these models cannot be applied to the NFH task directly, we propose a solution based on the model of \citet{lee2017end} which we adapt to incorporate the implicit cases.

\paragraph{Ellipsis} The most studied type of ellipsis is the Verb Phrase Ellipsis (VPE). Although the following refers to this line of studies, the task and resemblance to the NFH task hold up to the other types of ellipsis as well (e.g. Gapping \cite{lakoff1970gapping}, Sluicing \cite{john1969guess}, Nominal Ellipsis \cite{lobeck1995ellipsis}, etc.). VPE is the anaphoric process where a verbal constituent is partially or totally unexpressed but can be resolved through an antecedent from context \cite{liu2016exploring}. For example, in the sentence: ``His wife also \textit{works for the paper}, as \textbf{did} his father'', the verb \textbf{did} is used to represent the verb phrase \textit{works for the paper}.
The VPE resolution task is to detect the target word which creates the ellipsis and the anaphoric verb phrase which it depicts. Recent work 
\cite{liu2016exploring,kenyon2016verb} tackled this problem by dividing it into
two main parts: target detection and antecedent identification.

\paragraph{Semantic Graph Representations}
Several semantic graph representation cover some of the cases we consider.
Abstract Meaning Representation (AMR) is a graph-based semantic representation for language
\cite{pareja2013proceedings}. It covers a wide range of concepts and relations. Five of those concepts: \textit{year}, \textit{age}, \textit{monetary-quantity}, \textit{time} and \textit{person} correlate to our implicit classes: \textsc{Year}, \textsc{Age}, \textsc{Currency}, \textsc{Time} and \textsc{People} respectively.

The UCCA semantic representation \cite{abend-rappoport:2013:ACL2013} explicitly marks missing
information, including the \textsc{Reference} NFH cases, but not the \textsc{Implicit} ones.

\section{Conclusions}
Empty elements are pervasive in text, yet do not receive much research attention. In this work, we tackle a common phenomena that did not receive previous treatment.
We introduce the FH identification and resolution tasks and focus on a common and important FH subtype: the NFH. We demonstrate that the NFH is a common phenomenon, covering over 40\% of the number appearances in a large dialog-based corpus and a substantial amount in other corpora as well (>20\%). We create datasets for the NFH identification
and resolution tasks. We provide an accurate method for identifying the NFH
constructions and a neural baseline for the resolution task. The resolution task
proves challenging, requiring further research. We make the code and datasets
available to facilitate such research (\url{github.com/yanaiela/num_fh}).

\section*{Acknowledgments}
We would like to thank Reut Tsarfaty and the BIU NLP lab for the fruitful conversation and helpful comments.
The work was supported by 
The Israeli Science Foundation (grant number 1555/15) and 
the German Research Foundation via the
German-Israeli Project Cooperation (DIP, grant DA 1600/1-1).

\bibliography{tacl2018v2-template}
\bibliographystyle{acl_natbib}

\newpage
\appendix

\section{Details of Linear Baseline Implementation}
\label{sec:appendix-linear}
This section lists the features used for the linear baseline mentioned in Section \ref{sec:experiments-results}. The features are presented in Table \ref{tbl:linear-features}. We used four type of features: (1) \textbf{Label} features, making use of parsing labels of dependency and POS-taggers, as well as simple lexical features of the \textit{anchor}'s window. (2) \textbf{Structure} features, incorporating structural information from the sentence and the \textit{anchor}'s spans. (3) \textbf{Match} features test for specific patterns in the text, and (4) \textbf{Other}, not-categorized features.

We used the features described above to train a linear SVM classifier on the same splits. 
\begin{table}[h]
\centering
\resizebox{\columnwidth}{!}{%
\begin{tabular}{cl}
Type                                       & Feature Description                                                                    \\ \hline
\multirow{9}{*}{Labels}                    & Anchor \& Head Lemma                                                                           \\
                                           & 2 sized window lemmas                                                                  \\
                                           & 2 sized window POS tags                                                                \\
                                           & Dependency edge of target                                                              \\
                                           & Head POS tag                                                                           \\
                                           & Head lemma                                                                             \\
                                           & Left most child lemma of anchor head                                                   \\
                                           & Children of syntactic head                                                             \\ \hline
\multirow{8}{*}{Structure}                 & Question mark before or after the anchor                                               \\
                                           & Sentence length bin (\textless5\textless10\textless)                                   \\
                                           & Span length bin (1, 2 or more)                                                         \\
                                           & Hyphen in anchor span                                                                  \\
                                           & Slash in anchor span                                                                   \\
                                           & Apostrophe before or after the span                                                    \\
                                           & Apostrophe + 's' after span                                                            \\
                                           & Anchor is ending the sentence                                                          \\ \hline
\multicolumn{1}{l}{\multirow{3}{*}{Match}} & Whether the text contains a currency expression                                        \\
\multicolumn{1}{l}{}                       & Whether the text contains a time expression                                            \\
\multicolumn{1}{l}{}                       & Entity exists in the sentence before the target                                        \\ \hline
\multicolumn{1}{l}{\multirow{2}{*}{Other}} & Target size bin (\textless1\textless10\textless100\textless1600\textless2100\textless) \\
\multicolumn{1}{l}{}                       & The number shape (digit or written text)                                              
\end{tabular}
}
\caption{Features used for linear classifier.}
\label{tbl:linear-features}
\end{table}



\end{document}